# Feasibility of Structuring Stress Documentation Using an Ontology-Guided Large Language Model

Hyeoneui Kim[1,2,3*], Jeongha Kim[1,3], Huijing Xu[1], Jinsun Jung[1,2], Sunghoon Kang[4], Sun Joo Jang[1,2]

[1]College of Nursing, Seoul National University, Seoul, Republic of Korea

[2]The Research Institute of Nursing Science, Seoul National University, Seoul, Republic of Korea

[3]Center of Human-Caring Nurse Leaders for the Future by the Brain Korea (BK 21) four project, College of Nursing, Seoul National University, Seoul, Republic of Korea

[4]College of Medicine, Seoul National University, Seoul, Republic of Korea

* Corresponding Author

Hyeoneui Kim, RN, MPH, PhD

103 Daehak-ro, Jongno-gu, Seoul, 03080, Republic of Korea

ORCID ID: https://orcid.org/0000-0002-5931-7286

Email: ifilgood@snu.ac.kr

Phone: +82) 02-740-8483

**Abstract**

**Background**: Stress, arising from the dynamic interaction between external stressors, individual appraisals, and physiological or psychological responses, significantly impacts health yet is often underreported and inconsistently documented, typically captured as unstructured free-text in electronic health records. Ambient AI technologies offer promise in reducing documentation burden, but predominantly generate unstructured narratives, limiting downstream clinical utility.

**Purpose**: This study aimed to develop an ontology for mental stress and evaluate the feasibility of using a Large Language Model (LLM) to extract ontology-guided stress-related information from narrative text.

**Methods**: The Mental Stress Ontology (MeSO) was developed by integrating theoretical models like the Transactional Model of Stress with concepts from 11 validated stress assessment tools. MeSO's structure and content were refined using OntOlogy Pitfall Scanner! and expert validation. Using MeSO, six categories of stress – related information-stressor, stress response, coping strategy, duration, onset, and temporal profile – were extracted from 35 Reddit posts using Claude Sonnet 4. Human reviewers evaluated accuracy and ontology coverage.

**Results**: The final ontology included 181 concepts across eight top-level classes. Of 220 extractable stress-related items, the LLM correctly identified 172 (78.2%), misclassified 27 (12.3%), and missed 21 (9.5%). All correctly extracted items were accurately mapped to MeSO, although 24 relevant concepts were not yet represented in the ontology.

**Conclusion**: This study demonstrates the feasibility of using an ontology-guided LLM for structured extraction of stress-related information, offering potential to enhance the consistency and utility of stress documentation in ambient AI systems. Future work should involve clinical dialogue data and comparison across LLMs.

**Introduction**

**Significance of mental stress in human health**

Stress is a complex series of events that involves the interaction between external stress-inducing events, individual cognitive appraisals, and subsequent physiological and psychological responses. Stressors may arise from various aspects of one's environment, including social interactions, economic hardships, or health-related issues. The perception of stress, or the individual's cognitive appraisal of a stressor, plays a crucial role in determining the intensity and nature of the stress response. This appraisal is influenced by numerous factors such as prior experiences, personality traits, and coping mechanisms, which can either mitigate or exacerbate the effects of stress. The physiological processes underlying the stress response are predominantly regulated by the hypothalamic-pituitary-adrenal (HPA) axis and the sympathetic nervous system (SNS)[1]. These systems facilitate the body's adaptation to perceived threats by activating the "fight-or-flight" response. However, prolonged activation of these physiological pathways may result in dysregulation within the nervous and endocrine systems, leading to adverse health outcomes such as cardiovascular issues, gastrointestinal disturbances, and immune system dysfunction[2–4]. In addition, chronic stress can contribute to emotional exhaustion, resulting in burnout and a reduced ability to manage everyday challenges[5]. Consequently, the interplay between stressors, cognitive appraisals, and physiological and psychological stress responses forms a complex cycle that significantly influences an individual's overall well-being and quality of life.

Patients may experience higher levels of stress due to the physical and psychological burdens associated with their health conditions[6,7]. Specifically, the process of cancer diagnosis and treatment is highly stressful, often leading to feelings of uncertainty, fear, and emotional distress[8–10]. Previous studies have found that psychological distress is highly prevalent among breast cancer patients, as the disease often affects a woman's identity, body image, and self-esteem[11–13]. Significantly, most breast cancer patients are middle-aged women engaged in critical life stages such as family formation, career advancement, and child-rearing, which amplifies the psychosocial impact of a cancer diagnosis. These changes can negatively affect patients' adherence to treatment, complicating recovery prognosis[14,15]. For breast cancer patients, high stress levels not only impact disease progression and quality of life but also increase the risk of recurrence and reduce survival rates[16,17].

In addition, long-term hemodialysis patients with end-stage renal disease face significant psychological distress alongside treatment challenges[18,19]. The demanding nature of dialysis treatments, which often require frequent visits and prolonged sessions, coupled with strict dietary and fluid restrictions, leads to substantial lifestyle disruptions[20]. The physical discomfort and fatigue associated with these treatments further exacerbate patients' stress levels. The psychological distress caused by these circumstances may contribute to symptoms such as depression, anxiety, and sleep disturbances[21]. Given these complexities, healthcare providers are strategically positioned to identify and address the stressors of patients[6]. Therefore, if a patient's stress index and related information are identified early during the diagnosis and treatment processes, healthcare providers can implement stress management interventions that may alleviate stress and disrupt this vicious cycle in a timely manner[8,12,14].

**Challenges in documenting patient's stress experience in clinical settings**

Despite the significant impact of stress on health, patient's stress experiences are underreported in clinical settings[22]. Electronic health record (EHR) serves as a comprehensive repository of longitudinal patient data, systematically gathered by healthcare providers. Current clinical documentation practices in EHR predominantly focus on acute medical conditions and their immediate treatments, often lack structured frameworks for capturing and categorizing stress-related data[22]. Although many instruments are available for evaluating stress, there is no clear guidance on using appropriate stress assessment tools in clinical practice. For instance, general stress assessment tools such as 'Perceived Stress Scale (PSS)' and the stress subscale of the 'Depression, Anxiety, and Stress Scale (DASS)' are typically used to demonstrate wide-ranging effects of stress in both healthy and patient populations[23,24]. However, in the clinical context, these tools are primarily limited to patients with severe psychiatric disorders or major depression in psychiatric units. These tools are not routinely integrated into care plans for those with chronic illnesses such as cancer, cardiovascular diseases, or diabetes. Additionally, their use is at the discretion of the physician, limiting comprehensive stress evaluation across all patient groups. This gap persists despite evidence showing that life stressors can affect biological processes that contribute to the onset and progression of chronic conditions, including diabetes, obesity, and cardiovascular disease[25–28].

Stress-related information in EHRs is often documented through physician interviews and embedded in free-text narratives[29–31]. This lack of standardized formats makes systematic analysis challenging and may result in an incomplete understanding of the patient's stress profile. The absence of standardized documentation not only complicates the aggregation and analysis of patient health data, but also impedes effective communication among healthcare providers. For example, without a consistent way to categorize stress, subsequent healthcare providers may overlook important information that could affect treatment decisions, continuity of care, and individualized stress management strategies. This issue highlights the need for the systematic integration of comprehensive psychosocial assessments into routine healthcare practice to ensure a full understanding of factors affecting patient well-being.

**Generative AI assisted nursing documentation**

Recent advancements of generative AI have begun to impact various clinical domains. One of the most notable applications is ambient AI documentation, in which AI systems automatically capture patient-clinician conversations and convert them into clinical notes within EHR[32]. Studies have reported that ambient AI documentation systems can reduce documentation time and alleviate clinician frustration associated with EHR interaction[33,34]. Initially adopted for physicians, this technology is now being piloted for nursing workflows as well. However, current ambient AI systems primarily generate unstructured free-text notes, necessitating additional steps to extract key information for effective use of the documented information in patient care.

**Ontology-based solutions for addressing documentation challenges**

Ontologies are widely used to identify, manage, and share semantic knowledge within a defined domain, enhancing knowledge management and reasoning through structured data representation[35,36]. In a knowledge domain, concepts, attributes, and the relationships can be expressed through statements formulated in accordance with logic-based framework defined in ontologies[37]. This formalization promotes semantic interoperability across various systems and data sources. By leveraging ontology-based approaches, healthcare providers can consistently capture stress-related information, which is essential for understanding and addressing the multi-faceted impact on health. Considering that adaptive responses to stressors can promote both physiological and psychological resilience, it is essential to identify primary stressors and implement strategic stress management interventions that are informed by ontological frameworks. The Human Stress Ontology (HSO)[38] contains human stress concepts, including stress measurements, causes, effects, mediators and treatments. Nevertheless, the HSO exhibits limitations in complexity and accuracy, making it challenging to fully capture the multi-faceted nature of stress in clinical settings. In particular, the HSO does not encompass all potential concepts associated with the causes and effects of stress. Additionally, the Mental Health Ontology (MHO)[39] primarily focuses on mental disorders, utilizing diagnostic frameworks such as the 'International Classification of Diseases (ICD)' and the 'Diagnostic and Statistical Manual of Mental Disorders (DSM)'. Consequently, its applicability to general stress concepts is somewhat limited.

Furthermore, there is a lack of an appropriate ontology that effectively represents a person's stress experience. A stress ontology that captures the various dimensions of stress, including stressors, cognitive appraisals, physiological or psychological responses, and management treatments can facilitate the standardized documentation of stress in EHR. This approach ensures a holistic view of the patient's stress profile, ultimately leading to better health outcomes and more comprehensive patient care[14,40].

This study aimed to develop a stress ontology and evaluate the feasibility of using a Large Language Model (LLM), guided by the ontology, to automatically extract key stress-related information from narrative descriptions, in order to support structured stress documentation through ambient AI technologies

**Methods**

**Development of Mental Stress Ontology (MeSO)**

*Concept hierarchy construction*

We first constructed the backbone structure of MeSO benchmarking the top layer structure of Human Stress Ontology (HSO)[38] and the theoretical model of 'Lazarus and Folkman's 'Transactional Model of Stress (TMS)[41]'. The top-level classes of HSO offered a workable framework to categorize the key components of stress assessment such as 'stress measurements', 'stress causes', 'mediators', 'stress effects', and 'stress treatments'[38]. The concepts related to 'stress appraisal' and 'stress coping', which were crucial in planning nursing care on stress management, were adopted from TMS[42]. Stress appraisal is the process of determining if activating the stress response is needed or not. For example, if the stressor is threatening to the physical, psychological, social, and overall well-being of the individual (i.e., appraisal), the person is more likely to perceive the situation as stressful and activates coping strategies to manage stress[41].

To prepare the concepts to populate under the top-level classes, 266 questions from eleven stress assessment questionnaires listed in Table 1 were analyzed. Two reviewers manually extracted stress related keywords from these questions. The extracted keywords clustered based on conceptual similarities, and refined into distinct concept classes. These concept classes were then aggregated into broader concept groups, which formed higher-level (i.e., parent) concepts. This process continued until the top-level concepts were reached, resulting in a hierarchical structure.

The concept classes were labeled in English in singular nouns or gerunds taking the UpperCamelCase form, where each word's first letter capitalized. The Protégé (version 5.6.1)[42,43] was utilized to develop MeSO. Protégé is a widely used open-source ontology authoring and editing tool developed by Stanford University, providing a robust platform for constructing, managing, and visualizing ontologies[42].

**Table 1.** Stress assessment questionnaires used for ontology development.

| No. | Source |
|---|---|
| 1 | Coping Resources Inventory for Stress (CRIS)[44] |
| 2 | Depression Anxiety Stress Scales-21 (DASS-21)[24] |
| 3 | International Stress Management Association (ISMA) UK[45] |
| 4 | Perceived Stress Scale (PSS)[23] |
| 5 | Stress and Adversity Inventory for Adults (STRAIN)[46] |
| 6 | Stress Appraisal Measure (SAM)[47] |
| 7 | Stress Level Test (Self-Assessment)[48] |
| 8 | The Ardell Wellness Stress Test[49] |
| 9 | The Brief COPE Inventory (Carver)[50] |
| 10 | The Holmes and Rahe Social Readjustment Rating Scale (SRRS)[51] |
| 11 | The Work Stress Questionnaire (WSQ)[52] |

*Class annotation*

The concept classes in MeSO were annotated with the STRONG concept identifier, an internal identifier assigned by the authors, the Concept Unique Identifier (CUI), concept preferred name, and semantic type of the Unified Medical Language System (UMLS) Metathesaurus[53], a Korean translation of the concepts, language of the source questionnaire. The UMLS Metathesaurus is an extensive biomedical thesaurus developed by the National Library of Medicine, integrating numerous health and biomedical vocabularies to facilitate interoperability among diverse health information systems[53]. The authors conducted concept mapping to the UMLS Methathesaurus.

**Evaluation of MeSO**

MeSO was evaluated by checking concept coverage, structural quality, and consulting a mental health expert.

*Concept coverage evaluation using BERT*

Bidirectional Encoder Representations from Transformers (BERT), one of the pioneering LLM introduced by Google[54], was used to evaluate whether MeSO sufficiently covers important stress-related concepts.

A total of 58 stress-related texts, including the abstracts of scientific papers and online articles were collected. The important stress-related terms were extracted using KeyBERT. KeyBERT is a keyword extraction model that uses BERT to identify the most relevant keywords in a document[55]. Keyword extraction with KeyBERT takes several steps: KeyBERT creates a vector representation (i.e., embedding) of a document using BERT. KeyBERT generates candidate keywords from the document using N-gram[55], where users can determine the length of N-gram (i.e., unigram, bigram, trigram, etc.). Then KeyBERT converts the candidate keywords into an embedding using the same BERT model created with the document. Finally, KeyBERT calculates cosine similarity between each candidate keyword embedding and the document embedding, identifying the keywords with higher similarity scores. Ten keywords with the highest cosine similarity scores were identified for each text source at the unigram, bigram, and trigram levels. Bigram and trigram were employed to capture multi-word keywords.

The authors reviewed the extracted keywords and phrases after removing duplicates. A total of 82 unique keywords were obtained after normalizing lexical variations such as synonyms and pleural forms. These keywords were then mapped to the concepts in the stress ontology. The mapping results were marked with one of five categories based on semantic equivalence level: exact match – keywords have the concepts with the same meaning in MeSO; broader match – keywords can only relate to more general concepts in MeSO, meaning the former carry more specific meaning than the latter; narrower match – keywords encompass multiple concepts in MeSO, meaning the former carry more general meaning than the latter; partial equivalence – keywords and concepts in MeSO are related but non-hierarchically; no match – keywords have no related concepts in stress ontology.

*Structural quality evaluation*

The overall structural quality of MeSO was evaluated using the Ontology Debugger plug-in within Protégé[56] and OntOlogy Pitfall Scanner! (OOPS!)[57]. The Ontology Debugger plug-in is designed to evaluate the consistency and coherence of ontologies by checking the logical soundness of the concept structure[56]. OOPS! is a web-based application designed to automatically detect common errors in class naming and annotation as well as the structural issues in an ontology[57].

*Expert feedback*

The stress ontology was updated based on the content coverage and quality evaluation results. Then the updated ontology was re-evaluated by a mental health expert. The expert was asked to provide feedback on whether MeSO adequately captures the concept of human stress in a clear and non-overlapping manner, and whether the hierarchical relation among classes were appropriate.

**Extracting key stress-related information from free-text description**

To evaluate the feasibility of extracting structured stress-related information using LLMs and MeSO, we used a publicly available dataset of Reddit posts on stress experiences (N=2744) obtained from Kaggle. A random sample of 40 posts was selected for testing. Five posts were used to develop the prompt and evaluate the performance of three LLMs-Qwen3, Claude Sonnet 4, and GPT-4o. The LLMs were instructed to extract six key elements: stressor, stress response, stress coping strategy, stress duration, stress onset, and temporal profile of stress, and mapped them to corresponding concepts in MeSO. The descriptions of these information categories are summarized in Table 2. Zero-shot prompting was employed. Based on the initial performance testing, Claude Sonnet 4 was selected for further analysis. The remaining 35 posts were processed, and the extracted information and ontology mapping results were independently reviewed by two reviewers. The prompt used for information extraction is provided in Supplemental Table 1. The performance comparison results of the three LLMs are available at https://bit.ly/MeSOpretest.

**Table 2.** Stress-related information categories.

| Category | Description |
| --- | --- |
| Stressor | Source or causes of stress |
| Stress Response | Mental, emotional, physical, or behavioral reaction to stress |
| Stress Coping Strategy | The methods used to manage stress |
| Stress Duration | How long the stress lasts |
| Stress Onset | The manner in which stress begins – sudden or gradual |
| Stress Temporal Profile | The overall pattern of stress - acute or chronic |

**Results**

**Evaluation of MeSO**

The initial version of MeSO comprised 102 unique concepts organized under eight top-level classes representing key dimensions of stress: Stressor, Stress Mediator, Stress Appraisal, Stress Effect, and Stress Treatment.

To assess concept coverage, 82 newly prepared stress-related keywords were mapped to the stress ontology. Of these, 42 keywords (51.2%) had exact matches, and 34 (41.5%) had broader matches. Two keywords were mapped to more specific concepts (i.e., narrower matches), while four had no matches in the stress ontology. A complete list of the keywords and their mapping results is provided in Supplemental Table 2.

The 40 keywords without exact matches were reviewed again by the authors to determine their potential inclusion to the stress ontology. Following this review, 22 new concepts were added to MeSO. The ontology structure was also revised, expanding from five to eight top-level classes: Stressors, Stress Mediator, Stress Appraisal, Stress Response, Stress Intervention, Stress Coping Strategy, Stress Coping Outcome, and Stress Characteristics.

Structural validation using the Protégé Debugger plug-in confirmed that MeSO has a consistent and coherent structure. Further evaluation using the OntOlogy Pitfall Scanner! (OOPS!) identified minor issues, such as missing class definition and the absence of inverse property relationships. Since MeSO does not include defined classes, property relationships were not specified.

OOPS! also suggested that the concepts 'Restlessness' and 'Impatience' could be equivalent; however, this suggestion was not accepted, as the two concepts, though related, denote distinct phenomena – 'Restlessness' is typically a physical manifestation of mental discomfort, such as 'Impatience'.

A mental health expert confirmed that the stress ontology encompasses key concepts necessary for describing mental stress. While the ontology may not yet capture all relevant concepts, it was judged to have a solid conceptual framework that allows for future expansion as needed.

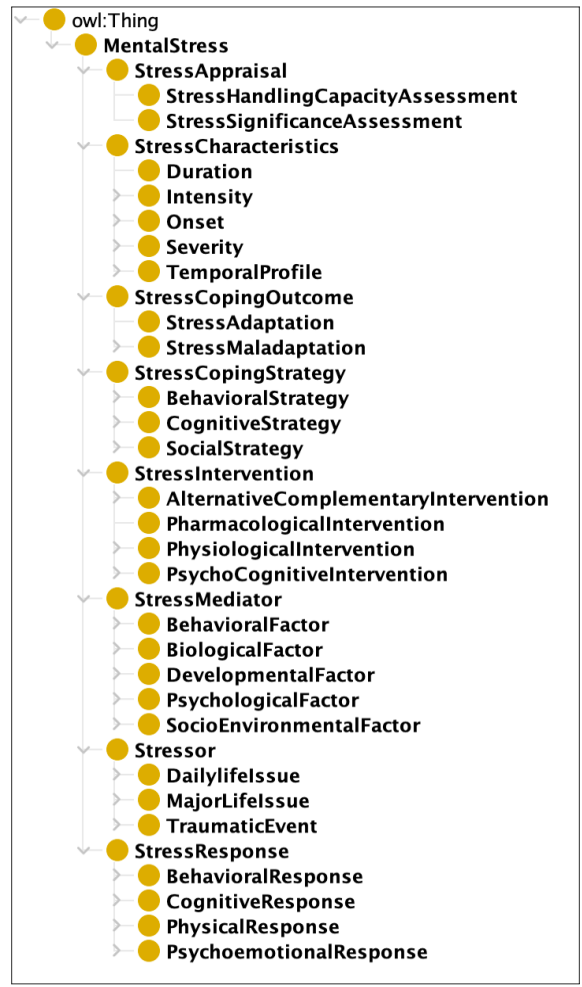

**Figure 1.** Top-level hierarchy of MeSO.

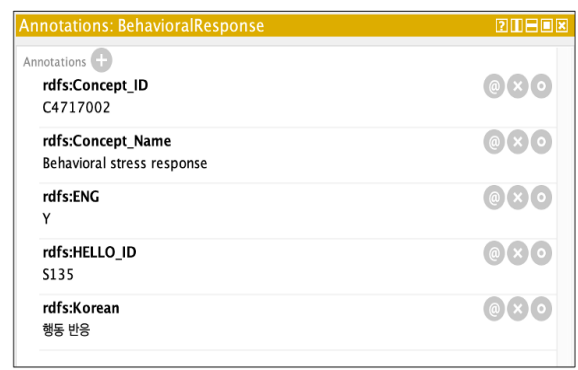

**Figure 2.** Class annotation example.

**Final version of MeSO**

The finalized version of MeSO included 181 concepts structured under eight top level classes, as shown in Figure 1. An example of the class annotation was also presented in Figure 2. MeSO is publicly available in BioPortal[58].

**Stress information extraction and ontology mapping**

The evaluation of extracted stress-related information showed a high level of inter-rater agreement between the two reviewers, with a weighted kappa score of 0.899. Discrepancies between the two primary reviewers were resolved through discussion with a third reviewer.

The performance of Claude Sonnet 4 in extracting stress-related information is summarized in Table 3. Many stress postings contained multiple stressors, stress responses, and stress coping strategies. Some postings, however, lacked extractable information – for example, two did not specify the cause of stress, and 21 did not mention the onset of stress.

Sonnet 4 accurately extracted information related to Stressor, Stress Response, Stress Duration, and Temporal Profile of Stress. However, it showed lower accuracy in identifying Stress Coping Strategy and Stress Onset. Most errors were attributed to missed cases (i.e., failures to detect existing information) or misinterpretations of the content. Three errors were identified as likely resulting from hallucination. Overall, there were total 220 stress related information items were identified as extractable by human reviewers. Of these, 78.2% were correctly extracted, 12.3% were incorrectly extracted due to misinterpretation, and 9.5% were missed entirely by Sonnet 4. The complete results – including the 35 stress postings, extracted information, and human evaluation – are provided in Supplemental Table 3.

**Table 3.** Performance of information extraction.

| Information Category | Correct, n (%) | Incorrect, n(%) | Missed, n (%) | Row Total |
|---|---|---|---|---|
| Stressor | 39 (81.25) | 0 (0.00) | 9 (18.75) | 48 |
| Stress Response | 55 (77.46) | 8 (11.27) | 8 (11.27) | 71 |
| Stress Coping Strategy | 21 (60.00) | 11 (31.43)[a] | 3 (8.57) | 35 |
| Stress Duration | 19 (86.36) | 3 (13.64) | 0 (0.00) | 22 |
| Stress Onset | 9 (64.29) | 5 (35.71)[b] | 0 (0.00) | 14 |
| Stress Temporal Profile | 29 (96.67) | 0 (0.00) | 1 (3.33) | 30 |
| **Column Total** | 172 (78.18) | 27 (12.27) | 21 (9.56) | 220 |

[a] one was caused by hallucination; [b] two were caused by hallucination

Claude Sonnet 4 accurately mapped extracted information to MeSO when a corresponding concept was available. However, fifty-two items could not be mapped. Among these, 22 were related to Stress Duration, which involved numeric values (e.g., number of years, months, and days) and were thus considered outside the scope of ontology mapping. Of the remaining 30 unmapped items, 20 were Stressor concepts, 6 were Stress Response concepts, and 4 were Stress Coping Strategy concepts. After removing duplicates, 16 unique unmapped Stressor concepts, 4 unique Stress Response concepts and 4 Stress Coping Strategy concepts remained. The unmapped concepts included vague expressions such as "unspecific stressors" and "multiple unspecific stressors," as well as more specific concepts like

"political difference," "career failure," and "lack of motivation". The complete list of unmapped concepts is provided in Supplemental Table 4.

**Discussion**

This study developed a mental stress ontology (MeSO) to support consistent and unambiguous representation of stress-related data. The ontology was then applied to guide Claude Sonnet 4 in automatically extracting key stress-related information from narrative descriptions of stress experience. This approach was designed to test the feasibility of capturing and documenting stress-related information in a structured format, with potential applicability in patient-provider interaction.

Given the reliability of modern speech-to-text technologies, this study began with written Reddit postings describing stress experiences, simulating transcribed patient speech. When guided by MeSO, Claude Sonnet 4 achieved an overall extraction accuracy of 78%. Furthermore, MeSO successfully covered 74% of the extractable concepts, underscoring its value in supporting standardized, structured representation of stress data.

Claude Sonnet 4 demonstrated high accuracy in extracting Stressors, Stress Response, and the Temporal Profile and Duration of stress. For example, it correctly identified 39 out of 48 extractable Stressor items. Among the nine missed cases, six involved a failure to recognize existing illness as a stressor. Stress Response extraction yielded a higher error rate (22.5%), primarily due to overinterpretation. For example, Claude Sonnet 4 interpreted expressions of self-harm ideation as indicative of depression rather than frustration, and a desire to stay in bed as social withdrawal rather than a lack of motivation. Claude Sonnet 4 also fails to synthesize emotional and behavioral cues into a coherent response concept.

On the other hand, Claude Sonnet 4 demonstrated greater difficulty in accurately identifying Stress Coping Strategy and Stress Onset. Within the coping strategy category, one error was classified as a hallucination, as no supporting evidence for the extracted information could be found in the original text. A substantial number of errors (6 out of 14) stemmed from misinterpreting venting behaviors in Reddit posts as intentional help-seeking. In this study, the research team defined help-seeking as a proactive behavior, typically involving explicit requests for advice (e.g., "Has anyone had a similar experience?" or "Does anyone have suggestions for dealing with this?"). Although refining the prompt to emphasize this distinction could potentially reduce such errors, providing overly detailed instructions would compromise scalability of the information extraction process. Conversely, Claude Sonnet 4 failed to recognize legitimate help-seeking behavior in two cases, suggesting limitations in detecting implicit or context-dependent expressions of coping strategies.

Regarding Stress Onset, Claude Sonnet 4 extracted onset information from 14 postings; however, 5 of them were incorrect. Two errors were due to unsupported inferences, representing hallucinations, while three resulted from misinterpreting gradual onset as sudden. Accurately identifying onset information often requires nuanced contextual inference, which remains a challenge for current LLMs. All hallucination cases are documented in Supplemental Table 3.

MeSO covered 74% of the extractable information (excluding Duration, which was deemed not mappable to the ontology due to its numeric nature), supporting its role as a standardized framework for representing stress-related information. The remaining unmapped concepts will be reviewed for possible inclusion in future version of MeSO, with refinement of expressed terms as needed.

Given the inherently subjective nature of mental stress, this study did not attempt to extract stress intensity. Moreover, even objectively definable information occasionally showed discrepancies between the LLM output and human reviewers. These findings underscore the continued need for clinician oversight when using AI-assisted documentation tools, particularly for nuanced or context-sensitive information. While AI can serve as a valuable support tool for clinical documentation, it is best viewed as an augmentation – not a replacement – of clinical judgement and expertise.

This study demonstrated the potential of combining a stress ontology with LLMs to support structured and scalable documentation of stress-related information from narrative input. By automating the extraction of key stress-related data such as stressors, stress responses, coping strategies, etc., this approach can reduce clinicians' documentation burden, improve the quality and consistency of stress documentation, and enhance interdisciplinary communication. This approach may also facilitate early identification of individuals experiencing significant stress, inform timely interventions, and lay the groundwork for future integration into ambient AI documentation tools in mental health settings.

**Limitations**

This study was conducted as a feasibility assessment of structured stress documentation using ambient AI technologies. However, it relied on written Reddit posts rather than real-world clinical conversations. While modern speech-to-text systems are generally accurate, real-world clinical environments may introduce background noise, diverse speech patterns, and contextual complexity that could impact both transcription quality and LLM performance. Moreover, stress-related content in clinical encounters is often embedded within broader health discussions, further complicating accurate information extraction.

In addition, only one LLM (Claude Sonnet 4) was used for information extraction, selected based on a preliminary evaluation involving five sample postings and a comparison with GPT and Qwen. Also, the primary feasibility test on information extraction was conducted on 35 postings. This small scale of evaluation limits generalizability and may not reflect the relative strengths of each LLM. Future research should include larger datasets, incorporate real clinical dialogue, and conduct systematic comparisons across multiple models including on-premises LLMs, especially given the privacy concerns surrounding clinical data.

**Conclusion**

Mental stress is a multifaceted phenomenon that can significantly influence physical and psychological health. This study developed a mental stress ontology and demonstrated its utility in guiding the automated extraction of key stress-related information from narrative text using a large language model. The results support the feasibility of using ontology-guided ambient AI technologies to produce structured and standardized documentation of stress-related data. This approach shows promises for improving the consistency, reusability, and interpretability of patient-reported stress information in clinical settings. As this approach is further validated and adopted, it can be expanded to include additional dimensions of mental stress, enhancing its utility across diverse care setting.

**Acknowledgements**

The authors have no acknowledgements to declare.

**Author contributions**

HK and SJ contributed to the conceptualization of the study. HK, JK and SK developed MeSO, conducted data analysis and drafted the manuscript. JJ, JK, and HX prepared the data and evaluated the LLM's performance. All authors contributed to the refinement and/or critical review of the manuscript. All authors agree with the final content of the manuscript.

**Declaration of conflicting interest**

The authors declared no potential conflicts of interest with respect to the research, authorship, and/or publication of this article.

**Ethical approval**

This study was exempt from the Institutional Review Board (IRB) of Seoul National University because it involved the analysis of publicly available de-identified secondary data (IRB No. E2507/004-005).

**Funding**

This study was supported in part by a grant from the National Research Foundation of Korea (grant number: 2022R1A2C201136011) and the BK21 four project (Center for Human-Caring Nurse Leaders for the Future) funded by the Ministry of Education and National Research Foundation of Korea.

**Data Availability**

The data used in this study are openly available at https://www.kaggle.com/datasets/mexwell/stress-detection-from-social-media-articles. The data that supports the findings of this study are available in the supplementary material.